\documentclass[runningheads]{llncs}
\usepackage[svgnames]{xcolor}
\usepackage{epsfig}
\usepackage{epstopdf}
\usepackage{graphicx}
\usepackage{amsmath,amssymb}
\usepackage{color}
\usepackage{textcomp}
\usepackage{gensymb}
\usepackage{booktabs}
\usepackage{bbm}
\usepackage{bm}
\usepackage{floatrow}
\usepackage[nocompress]{cite}
\usepackage{enumitem}
\providecommand\KGfinal[1]{\textcolor{Black}{#1}}
\providecommand\changes[1]{\textcolor{Black}{#1}}
\providecommand\CRchanges[1]{\textcolor{Black}{#1}}
\providecommand\JD[1]{\textcolor{Black}{#1}}
\providecommand\JDtwo[1]{\textcolor{Black}{#1}}
\providecommand\JDthree[1]{\textcolor{Black}{#1}}
\providecommand\JDfour[1]{\textcolor{Black}{#1}}
\providecommand\KG[1]{\textcolor{Black}{#1}}
\providecommand\KGfour[1]{\textcolor{Black}{#1}}

\newcommand{\featurename}[0]{{ShapeCode}}

\begin{document}
\title{ShapeCodes: Self-Supervised Feature Learning\\by Lifting Views to Viewgrids}

\titlerunning{ShapeCodes: Self-Supervised Feature Learning by Lifting Views to Viewgrids}
\authorrunning{D. Jayaraman, R. Gao, and K. Grauman}

\author{Dinesh Jayaraman\inst{1,2} \and
Ruohan Gao\inst{2} \and
Kristen Grauman\inst{3,2}}

\institute{UC Berkeley \and
  UT Austin \and
  Facebook AI Research
}

\maketitle
\begin{abstract}
  \vspace{-0.1in}
We introduce an unsupervised feature learning approach that embeds 3D shape information into a single-view image representation.  The main idea is a self-supervised training objective that, given only a single 2D image, requires all unseen views of the object to be predictable from learned features.  We implement this idea as an encoder-decoder convolutional neural network.  The network maps an input image of an unknown category and unknown viewpoint to a latent space, from which a deconvolutional decoder can best ``lift'' the image to its complete viewgrid showing the object from all viewing angles.  %
Our class-agnostic training procedure encourages the representation to capture fundamental shape primitives and semantic regularities in a data-driven manner---without manual semantic labels.  Our results on two widely-used shape datasets show 1) our approach successfully learns to perform ``mental rotation'' even for objects unseen during training, and 2) the learned latent space is a powerful representation for object recognition, outperforming several existing unsupervised feature learning methods.

\end{abstract}
\section{Introduction}
\vspace{-0.05in}

The field has made tremendous progress on object recognition by learning image features from supervised image datasets labeled by object categories~\cite{alexnet,resnet,Girshick_2014_CVPR}.  Methods tackling today's challenging recognition benchmarks like ImageNet or COCO~\cite{imagenet,coco} capitalize on the 2D regularity of Web photos to discover useful appearance patterns and textures that distinguish many object categories.  However, there are limits to this formula: manual supervision is notoriously expensive, not all objects are well-defined by their texture, and (implicitly) learning viewpoint-specific category models is \KGfour{cumbersome} if not unscalable.

\begin{figure}[t]
  \floatbox[{\capbeside\thisfloatsetup{capbesideposition={right,top},capbesidewidth=0.5\linewidth}}]{figure}[\FBwidth]
  {\caption{\small{Learning \emph{ShapeCodes} by lifting views to ``viewgrids''.
  \KGfour{Given} one 2D view of an \KGfour{unseen} object \KGfour{(possibly from an unseen category)}, \KGfour{our deep network} learns to produce the remaining views in the viewgrid. This self-supervised learning induces a \KGfour{feature space for recognition.  It} embeds valuable cues about 3D shape regularities that transcend object category boundaries.}}\label{fig:one_shot_concept}}%
  {\includegraphics[width=1\linewidth]{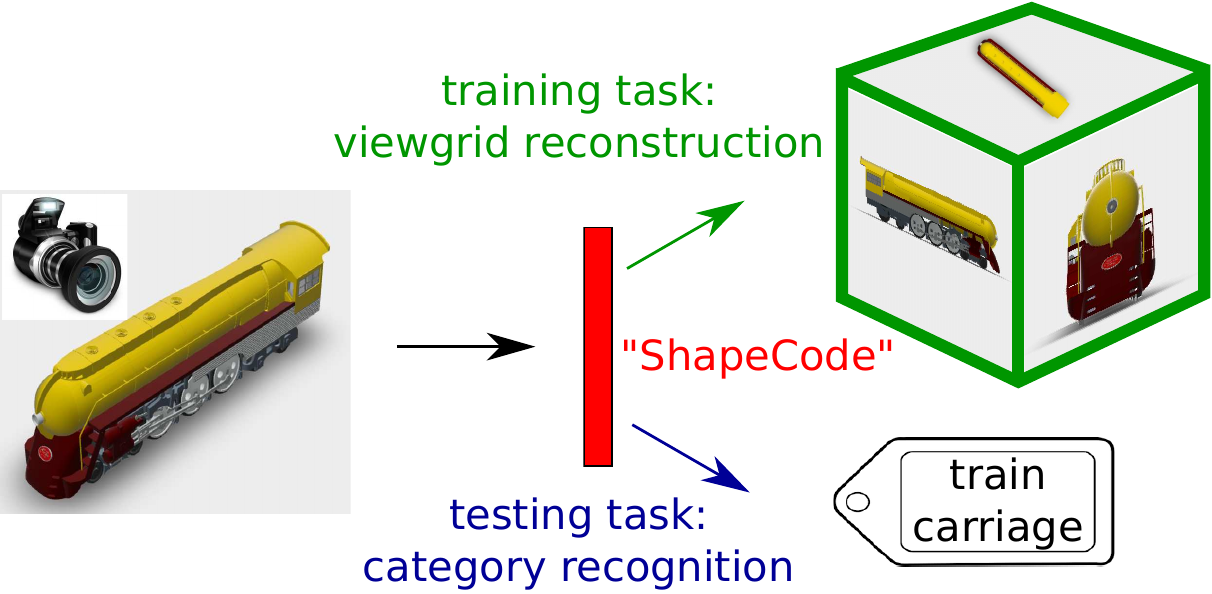}}
\vspace{-0.15in}
\end{figure}

Restricting learned representations to a 2D domain presents a fundamental handicap.
While visual perception relies largely on 2D observations, objects in the world are inherently three-dimensional entities.  In fact, cognitive psychologists find strong evidence that inferring 3D geometry from 2D views is a useful skill in human perception.  For example, in their seminal ``mental rotation" work, Shepard and colleagues observed that people attempting to determine if two views portray the same abstract 3D shape spend time that is linearly proportional to the 3D angular rotation between those views~\cite{shepard1971mental}. %
Such findings suggest that humans may explicitly construct mental representations of 3D shape from individual 2D views, and further, the act of mentally rotating such representations is integral to registering object views, and by extension, to object recognition.

Inspired by this premise, we propose an unsupervised approach to image feature learning that aims to ``lift'' individual 2D views to their 3D shapes.  More concretely, we pose the feature learning task as one-shot viewgrid prediction from a single input view.  A viewgrid---an array of views indexed by viewpoints---serves as an implicit image-based model of 3D shape.  We implement our idea as an encoder-decoder deep convolutional neural network (CNN).  Given one 2D view of any object from an arbitrary viewpoint, the proposed training objective learns \emph{a latent space from which images of the object after arbitrary rotations are predictable}.
See Fig~\ref{fig:one_shot_concept}.  Our approach extracts this latent ``\featurename'' encoding to generate image features for recognition.  Importantly, our approach is \emph{class-agnostic}: it learns a single model to accommodate all objects seen during training, thereby encouraging the representation to capture basic shape primitives, semantic regularities, and shading cues.  Furthermore, the approach is \emph{self-supervised}: it aims to learn a representation generically useful to object perception, but without manual semantic labels.

We envision the proposed training phase as if an embodied visual agent were learning visual perception from scratch by inspecting a large number of objects.  It can move its camera to arbitrary viewpoints around each object to acquire its own supervision.  At test time, it must be able to observe only one view and hallucinate the effects of all camera displacements from that view.  In doing so, it secures a representation of objects that, in a departure from purely 2D view-based recognition, is inherently shape-aware.  \KGfour{We discuss the advantages of viewgrids over explicit voxels/point clouds in Sec.~\ref{sec:task}.}

Our work relates to a growing body of work on self-supervised representation learning~\cite{efros-context,jigsaws,dinesh-iccv2015,lsm,shakh-colorization,pathak2016context,split-brain,gupta-video,ruohan-video,slow-steady}, where ``pretext'' tasks requiring no manual labels are posed to the feature learner to instill visual concepts useful for recognition.  We pursue a new dimension in this area---instilling 3D reasoning about shape.  The idea of inferring novel viewpoints also relates to work in CNN-based view synthesis and 3D reconstruction~\cite{ChairsCVPR2015,ZhouViewSynthesis2017,ChoySavarese2016,GirdharLearning2016,Renzede,YanPerspective2016,KarTulsianiCategorySpecific,Freeman-ECCV2016,TatarchenkoMultiview2016}.  However, our intent and assumptions are quite different.  Unlike our approach, prior work targets reconstruction as the end task itself (not recognition), builds category-specific models (e.g., \KGfour{one model for} chairs, \KGfour{another for} cars), and relies on networks pretrained with heavy manual supervision.

Our experiments on two widely used object/shape datasets validate that (1) our approach successfully learns class-agnostic image-based shape reconstruction, generalizing even to categories that were not seen during training and (2) the representations learned in the process \KGfinal{transfer} well to object recognition, outperforming several popular unsupervised feature learning approaches. Our results establish the promise of explicitly targeting 3D understanding as a means to learn \KGfour{useful} \KGfinal{image} representations.

\vspace{-0.1in}
\section{Related Work}
\vspace{-0.05in}

\paragraph{Unsupervised representation learning}  While supervised ``pretraining" of CNNs with large labeled datasets is useful~\cite{Girshick_2014_CVPR}, it comes at a high supervision cost and there are limits to its transferability to tasks unlike the original labeled categories.  As an increasingly competitive approach, researchers investigate \emph{unsupervised} feature learning~\cite{coates2011,efros-context,jigsaws,pathak2016context,dinesh-iccv2015,lsm,shakh-colorization,split-brain,gupta-video,ruohan-video,slow-steady,wang2017transitive,bojanowski2017unsupervised}.  An emerging theme is to identify ``pretext" tasks, where the network targets an objective for which supervision is inherently free.  In particular, features tasked with being predictive of contextual layout~\cite{efros-context,pathak2016context,jigsaws}, camera egomotion~\cite{dinesh-iccv2015,lsm,poier2018learning}, stereo disparities~\cite{gan2018geometry}, colorization~\cite{shakh-colorization}, or temporal slowness~\cite{gupta-video,ruohan-video,slow-steady} simultaneously embed basic visual concepts useful for recognition.  Our approach shares this self-supervised spirit and can be seen as a new way to force the visual learner to pick up on fundamental cues. In particular, our method expands this family of methods to multi-view 3D data\JD{, addressing the question: does learning to infer 3D from 2D help perform object recognition?}  While prior work considers 3D egomotion~\cite{dinesh-iccv2015,lsm}, it \JD{is restricted to impoverished glimpses of the 3D world through ``unattached'' neighboring view pairs from video sequences. Instead, our approach leverages viewgrid representations of full 3D object shape}. Our experiments comparing to egomotion self-supervision show our method's advantage.

%

%

\vspace*{-0.1in}
\paragraph{Recognition of 3D objects} Though 2D object models dominate recognition in recent years (e.g., as evidenced in challenges like PASCAL, COCO, ImageNet), there is growing interest in grounding object models in their 3D structures and shapes.  Recent contributions in large-scale data collections are fostering such progress~\cite{WuShapenets2015,ChangShapeNet,modelnet}, and researchers are developing models to integrate volumetric and multiview approaches effectively~\cite{QiVolumetric2016,SuMultiview2015}, as well as new ideas for relating 3D properties (pose, occlusion boundaries) to 2D recognition schemes~\cite{XiangDataDriven}.
Active recognition methods reason about the information value of unseen views of an object~\cite{schiele-transinformation,ramanathan-pinz2011,WuShapenets2015,JohnsCVPR2016,dinesh-eccv2016,dinesh-tpami2018}.

\vspace*{-0.1in}
\paragraph{Geometric view synthesis} For many years, new view synthesis was solved with geometry.  In image-based rendering, rather than explicitly constructing a 3D model,
new views are generated directly from multiple 2D views~\cite{Kang-survey}, with methods that establish correspondence and warp pixels according to projective or multi-view geometry~\cite{seitz-dyer-view-morphing,avidan-shashua-novel-view}.
Image-based models for object shape (implicitly) intersect silhouette images to carve a visual hull~\cite{matusik2000,space-carving}.%

\vspace*{-0.1in}
\paragraph{Learning 2D-3D relationships} More recently, there is interest in instead \emph{learning} the connection between a view and its underlying 3D shape.  The problem is tackled on two main fronts: image-based and volumetric.  Image-based methods infer the new view as a function of a specified viewpoint.
Given two 2D views, they learn to predict intermediate views~\cite{transforming-autoencoders,DeepStereo2015,DeepMorphing,DingMental2014}.
Given only a single view, they learn to render the observed object from new camera poses, e.g., via disentanglement with deep inverse graphic networks~\cite{KulkarniGraphicsNet2015}, tensor completion~\cite{ChaoYehPeople2014}, \KGfour{recurrent encoder-decoder nets~\cite{YangWeaklySupervised,lookaround-arxiv}}, \KGfour{appearance flow}~\cite{ZhouViewSynthesis2017}, \KGfour{or converting partial RGBD to panoramas~\cite{Im2Pano3D}}.  Access to synthetic object models is especially valuable to train a generative CNN~\cite{ChairsCVPR2015}.
Volumetric approaches instead map a view(s) directly to a 3D representation of the object, such as a voxel occupancy grid or a point cloud, \KGfour{e.g., with 3D recurrent networks~\cite{ChoySavarese2016}, direct prediction of 3D points~\cite{fan2017point}, or generative embeddings~\cite{GirdharLearning2016}.}
While most efforts study synthetic 3D object models (e.g., CAD datasets), recent work also ventures into real-world natural imagery~\cite{KarTulsianiCategorySpecific}.  Beyond voxels, inferring depth maps~\cite{TatarchenkoMultiview2016} or keypoint skeletons~\cite{Freeman-ECCV2016} offer valuable representations of 3D structure.

Our work builds on such advances in learning 2D-3D ties, and our particular convolutional autoencoder (CAE)-based pipeline (Sec.~\ref{sec:arch}) echoes the de facto standard architecture for pixel-output tasks~\cite{TatarchenkoMultiview2016,KulkarniGraphicsNet2015,masci2011stacked,YanPerspective2016,split-brain}.  However, our goal differs from any of the above.  Whereas existing methods develop category-specific models (e.g., chairs, cars, faces) and seek high-quality images/voxels as the end product, we train a class-agnostic model and seek a transferable image representation for recognition.
\section{Approach}
\vspace{-0.05in}

Our goal is to learn a representation that lifts a single image from an arbitrary (unknown) viewpoint and arbitrary class to a space where the object's 3D shape is predictable---its \emph{ShapeCode}.
This task of ``mentally rotating'' an object from its observed viewpoint to arbitrary relative poses requires \KGfour{3D} understanding from single 2D views, which is valuable for \KGfour{recognition}.
By training on a one-shot shape reconstruction task, our approach aims to learn an image representation that embeds this 3D understanding and applies the resulting embedding for single-view recognition tasks.

\subsection{Task setup: One-shot viewgrid prediction}\label{sec:task}

During training, we first evenly sample views from the viewing sphere around each object. To do this, we begin by selecting a set $S_{az}$ of $M$ camera azimuths $S_{az}=\{360\degree/M, 720\degree/M, \dots 360\degree\}$ centered around the object. Then we select a set $S_{el}$ of $N$ camera elevations $S_{el}=\{0\degree, \pm 180\degree/(N-1), \pm 360\degree/(N-1),\dots \pm 90\degree \}$. We now sample all $M\times N$ views of every object corresponding to the cartesian product $S=S_{az}\times S_{el}$ of azimuth and elevation positions: $\{\bm{y}(\bm{\theta_i}): \bm{\theta_i}\in S\}$.\footnote{omitting object indices throughout to simplify notation.} Note that each $\bm{\theta_i}$ is an elevation-azimuth pair, and represents one position in the viewing grid $S$.

Now, with these evenly sampled views, the one-shot viewgrid prediction task can be formulated as follows. Suppose the observed view is at an unknown camera position $\bm{\theta}$ sampled from our viewing grid set $S$ of camera positions. The system must learn to predict the views $\bm{y}(\bm{\theta'})$ at position $\bm{\theta'}=\bm{\theta}+\bm{\delta_i}$ for all $\bm{\delta_i}\in S$. Because of the even sampling over the full viewing sphere, $\bm{\theta}'$ is itself also in our original viewpoint set $S$, so we have already acquired supervision for all views that our system must learn to predict.

\KGfour{Why viewgrids?  The viewgrid representation has advantages over other more explicit 3D representations such as point clouds~\cite{fan2017point} and voxel grids~\cite{ChoySavarese2016}.  First, viewgrid \KGfinal{images} can be directly acquired by an embodied agent through object manipulation or inspection, whereas voxel grids and point clouds require noisy 3D inference from large image collections.  While our experiments leverage realistic 3D object CAD models to render viewpoints on demand (Sec.~\ref{sec:results}), it is actually less dependent on CAD data than prior work requiring voxel supervision.  Ultimately we envision training taking place in a physical scenario, where an embodied agent builds up its visual representation by examining various objects.  By moving to arbitrary viewpoints around an object, it acquires self-supervision to understand the 3D shape.  Finally, viewgrids faciliate the representation of missing data---if some ground truth views are unavailable for a particular object, the only change required in our training loss (below in  Eq~\ref{eq:loss}) would be to drop the terms corresponding to unseen views.}

\subsection{Network architecture and training}\label{sec:arch}

To tackle the one-shot viewgrid prediction task, we employ a deep feed-forward neural network.
Our network architecture naturally splits into four modular sub-networks with different functions: elevation sensor, image sensor, fusion, and finally, a decoder. Together, the elevation sensor, image sensor, and fusion modules process the observation and proprioceptive camera elevation information to produce a single feature vector that encodes the full object model. \JD{That vector space constitutes the learned \emph{ShapeCode} representation.}  During training \KGfour{\emph{only}}, the decoder module %
\KGfour{processes} this code through a series of learned deconvolutions to produce the desired image-based viewgrid reconstruction at its output.

\vspace*{-0.1in}
\paragraph{Encoder:} First, the image sensor module embeds the observed view through a series of convolutions and a fully connected layer into a vector. In parallel, the camera elevation angle is processed through the elevation sensor module. Note that the object pose is not fully known---while camera elevation can be determined from gravity cues, there is no way to determine the azimuth.

The outputs of the image and elevation sensor modules are concatenated and passed through a fusion module which jointly processes \KGfour{their} information to produce a $D=256$-dimensional output ``code'' vector, which embeds knowledge of object shape. \KGfour{In short,} the function of the encoder is to lift a 2D view to a single vector representation of the full 3D object shape.

\vspace*{-0.1in}
\paragraph{Decoder:} To learn a representation with this property, the output of the encoder is processed through another fully connected layer to increase its dimensionality before reshaping into a sequence of small $4\times 4$ feature maps. These maps are then iteratively upsampled through a series of learned deconvolutional layers. The final output of the decoder module is a sequence of $MN$ output maps $\{\bm{\hat{y}_i}: i=1,\dots M\times N\}$ of same height and width as the input image. Together these $MN$ maps represent the system's output viewgrid on which the training loss is computed. %

\begin{figure*}[t]
  \vspace{-0.2in}
  \centering
  \includegraphics[width=1\textwidth]{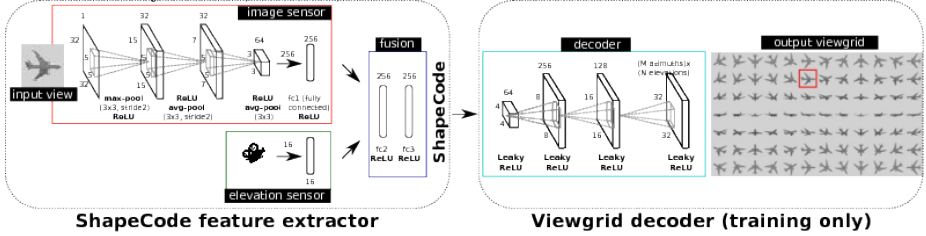}
  \caption{Architecture of our system. A single view of an object (top left) and the corresponding camera elevation (bottom left) are processed independently in ``image sensor'' and ``elevation sensor'' neural net modules, before fusion to produce the \emph{ShapeCode} representation of the input, which embeds the 3D object shape aligned to the observed view. This is now processed in a deconvolutional decoder. During training, the output is a sequence of images representing systematically shifted viewpoints relative to the observed view. \JD{During testing, novel 2D images are lifted into the ShapeCode representation to perform recognition.}}
  \label{fig:architecture}
  \vspace{-0.1in}
\end{figure*}

The complete architecture, together with more detailed specifications, is visualized in Fig~\ref{fig:architecture}.  Our convolutional encoder-decoder~\cite{masci2011stacked} neural network \KGfour{architecture is similar to~\cite{TatarchenkoMultiview2016,YanPerspective2016,ZhouViewSynthesis2017,lookaround-arxiv}.} As discussed above, however, the primary focus of our work is very different. We consider one-shot reconstruction as a path to useful image representations that lift 2D views to 3D, whereas existing work addresses the image/voxel generation task itself, and accordingly builds category-specific models.

By design, our approach must not exploit artificial knowledge about the absolute orientation of an object it inspects, either during training or testing.
Thus, there is an important issue to deal with---what is the correspondence between the individual views in the target viewgrid and the system's output maps? Recall that at test time, the system will be presented with a single view of a novel object, from an unknown viewpoint (elevation known, azimuth unknown). How then can it know the correct viewpoint coordinates for the viewgrid it must produce? It instead produces viewgrids aligned with the \emph{observed} viewpoint at the azimuthal coordinate origin, similar to~\cite{lookaround-arxiv}. Azimuthal rotations of a given viewgrid all form an equivalence class. In other words, circularly shifting the 7$\times$12 viewgrid in Fig~\ref{fig:architecture} by one column will produce a different, but entirely valid viewgrid representation of the same airplane object.

To optimize the entire pipeline, we regress to the target viewgrid $\bm{y}$, which is available for each training object. Since our output viewgrids are aligned by the observed view, we must accordingly shift the target viewgrid before performing regression. This leads to the following minimization objective:
  \vspace{-0.1in}
\begin{equation}
  \mathcal{L}=\sum_{i=1}^{M\times N} \|\bm{\hat{y}_i} - \bm{y}(\bm{\theta}+\bm{\delta_i})\|^2,
  \label{eq:loss}
  \vspace{-0.07in}
\end{equation}
where we omit the summation over the training set to keep the notation simple. Each output map $\bm{\hat{y}_i}$ is thus penalized for deviation from a specific \emph{relative} circular shift $\bm{\delta_i}$ from the observed viewpoint $\bm{\theta}$. A similar reconstruction loss was proposed in~\cite{lookaround-arxiv} in a different context; they learned exploratory action policies by training them to select a sequence of views best suited to reconstruct entire objects and scenes.


\JDtwo{Recent work targeting image synthesis has benefited from using adversarial (GAN) losses~\cite{gan}. GAN losses help achieve correct low-level statistics in image patches, improving photorealism~\cite{isola2016image}. Rather than targeting realistic image synthesis as an end in itself, we target shape reconstruction for \emph{feature} learning, so we use the standard $\ell_2$ loss. During feature transfer, we discard the decoder entirely (see Sec.~\ref{sec:nn_approach}.)}

When generating unseen views, our system recovers the full viewgrid in one shot, i.e., in a single forward pass through the neural network. %
Directly producing the full viewgrid has several advantages. In particular, it is efficient, and it avoids ``drift'' degradation from iterated forward passes. Most critically to our end goal of unsupervised feature learning, this one-shot reconstruction task enforces that the encoder must capture the \emph{full} 3D object shape from observing just one 2D view.

Models are initialized with parameters drawn from a uniform distribution between -0.1 and +0.1. The mean squared error loss is then optimized through standard minibatch stochastic gradient descent (batch size 32, momentum 0.9) via backpropagation. For our method and all baselines in Sec~\ref{sec:results}, we optimize the learning rate hyperparameter on validation data. Training terminates when the loss on the validation set begins to rise.

\subsection{\JD{ShapeCode features} for object recognition}~\label{sec:nn_approach}
\vspace{-0.1in}

\JD{During training, the objective is to minimize viewgrid error, to learn the latent space from which unseen views are predictable.  Then, to apply our network to novel examples, the representation of interest is that same latent space output by the fusion module of the encoder---the ShapeCode.}
In the spirit of self-supervised representation learning, we hypothesize that features trained in this manner will facilitate high-level visual recognition tasks. This is motivated by the fact that in order to solve the reconstruction task effectively, the network must implicitly learn to lift 2D views of objects to inferred 3D shapes. A full 3D shape representation has many attractive properties for generic visual tasks. \changes{For instance, pose invariance is desirable for recognition; while difficult in 2D views, it becomes trivial in a 3D representation, since different poses correspond to simple transformations in the 3D space.}  \KG{Furthermore, the ShapeCode provides a representation that is \emph{equivariant} to egomotion transformations, which is known to benefit recognition and emerges naturally in supervised networks~\cite{dinesh-ijcv2017,cohen2014transformation,lenc2015understanding,transforming-autoencoders}}.

Suppose that the visual agent has learned a model as above for viewgrid prediction by inspecting 3D shapes. Now it is presented with a new recognition task, as encapsulated by a dataset of class-labeled training images \KGfour{\emph{from a disjoint set of object categories}}.  We aim to transfer the 3D knowledge acquired in the one-view reconstruction task to this new task.
Specifically, for each new class-labeled image,  we directly represent it in the feature space represented by an intermediate fusion layer in the network trained for reconstruction. These features are then input to a generic machine learning pipeline that is trained for the categorization task.  %

\changes{Recall that the output of the fusion module in Fig~\ref{fig:architecture}, which is the fc3 feature vector, is trained to encode 3D shape. In our experiments, we test the usefulness of features from fc3 and its two immediate preceding layers, fc2, and fc1, for solving object classification and retrieval tasks.}
\section{Experiments}\label{sec:results}
\vspace{-0.05in}

First, we quantify performance for class-agnostic viewgrid completion \KGfour{(Sec~\ref{sec:exp_reconstr}).}  Second, we evaluate the learned features  for object recognition (\KGfour{Sec~\ref{sec:unsup}}).

\subsection{Datasets}

\JD{In principle, our self-supervised learning approach can leverage viewpoint-calibrated viewgrids acquired by an agent systematically inspecting objects in its environment. In our experiments, we generate such viewgrids from datasets of synthetic object shapes.} We test our method on two such publicly available datasets: ModelNet~\cite{modelnet} and ShapeNet~\cite{shapenet}. Both of these datasets provide a large number of manually generated 3D models, with class labels. For each object model, we render 32$\times$32 grayscale views from a grid of viewpoints that is evenly sampled over the viewing sphere centered on the object.

\textbf{ModelNet}~\cite{modelnet} has 3D CAD models downloaded from the Web, and then manually aligned and categorized. ModelNet comes with two standard subsets: ModelNet-10 and ModelNet-40, with 10 and 40 object classes respectively. The 40 classes in ModelNet-40 include the 10 classes in ModelNet-10. We use the 10 ModelNet-10 classes as our \emph{unseen classes}, and the other 30 ModelNet-40 classes as \emph{seen classes.} We use the standard train-test split, and set aside 20\% of seen-class test set models as validation data. \JDthree{ModelNet is the most widely used dataset in recent 3D object categorization work~\cite{JohnsCVPR2016,dinesh-eccv2016,dinesh-tpami2018,SuMultiview2015,modelnet,wu2016learning,qi2016volumetric}.}

\textbf{ShapeNet}~\cite{shapenet} contains a large number of models organized into semantic categories under the WordNet taxonomy. \changes{All models are consistently aligned to fixed canonical viewpoints.} We use the standard ShapeNetCore-v2 subset which contains 55 diverse categories. Of these, we select the 30 largest categories as seen categories, and the remaining 25 are unseen. We use the standard train-test split. Further, since different categories have highly varying numbers of object instances, we limit each category in our seen-class training set to 500 models, to prevent training from being dominated by models of a few very common categories. Table~\ref{tab:datasets} (left) shows more details for both datasets.

\vspace{-0.05in}
\subsection{Class-agnostic one-shot viewgrid prediction}~\label{sec:exp_reconstr}
\vspace{-0.15in}

First, we train and test our method \KGfour{for viewgrid prediction.} For both datasets, the system is trained on the seen-classes training set.
The trained model is subsequently tested on both seen and unseen class test sets.

The evaluation metric is the per-pixel mean squared deviation of the inferred viewgrid vs.~the ground truth viewgrid. We compare to several baselines: %
\begin{itemize}[leftmargin=*]
  \item \textbf{Avg view:} This baseline simply predicts, at each viewpoint in the viewgrid, the average of all views observed in the training set \emph{over all viewpoints}.
  \item \textbf{Avg viewgrid:} Both ModelNet and ShapeNet have consistently aligned models, so there are significant biases that can be exploited by a method that has access to this canonical alignment information. This baseline aims to exploit this bias by predicting, at each viewpoint in the viewgrid, the average of all views observed in the training set \emph{at that viewpoint}. Note that our system does \textbf{not} have access to this alignment information, so it cannot exploit this bias.
  \item \textbf{GT class avg view:} This baseline represents a model with perfect object classification. Given an arbitrary object from some ground truth category, this baseline predicts, at each viewpoint, the average of all views observed in the training set for that category.
  \item \textbf{GT class avg viewgrid:} This baseline is the same as GT category avg view, but has knowledge of canonical alignments too, so it produces the average of views observed at each viewpoint over all models in that category in the training set.
  \item \textbf{Ours w.~CA:} This baseline is our approach but trained with the (unrealistic) addition of knowledge about canonical alignment (``CA'') of viewgrids.  It replaces Eq~\eqref{eq:loss} to instead optimize the loss:
  $\mathcal{L}=\sum_{i=1}^{M\times N} (\bm{\hat{y}_i} - \bm{y}(\bm{\delta_i}))^2$,
so that each output map $\bm{\hat{y}_i}$ of the system is now assigned to a specific coordinate in the canonical viewgrid axes. %
\end{itemize}
A key question in these experiments is whether the class-agnostic model we train can generalize to predict unseen views of objects from \emph{classes} never seen during training.\footnote{\KG{For this very reason, it is not clear how to map existing view synthesis models that are category-specific (e.g., chairs in~\cite{ChairsCVPR2015}, chairs/cars in~\cite{ZhouViewSynthesis2017}) to our class-agnostic setting so that they could compete fairly.}}

\begin{table}[t] %
  \centering
  \resizebox{0.49\textwidth}{0.12\textwidth}{
  \small{
    \begin{tabular}{lcc|cc}
    \toprule
  Dataset$\rightarrow$                  & \multicolumn{2}{c|}{ModelNet} & \multicolumn{2}{c}{ShapeNet} \\
    Methods$\downarrow$/ Data$\rightarrow$ & \multicolumn{1}{c}{seen} & \multicolumn{1}{c|}{unseen} & \multicolumn{1}{c}{seen} & \multicolumn{1}{c}{unseen} \\\midrule
    Cam elevations   & \multicolumn{2}{c|}{0,$\pm$30\degree,$\pm$60\degree,$\pm$90\degree} & \multicolumn{2}{c}{0,$\pm$30\degree,$\pm$60\degree} \\
    Cam azimuths     & \multicolumn{2}{c|}{0,30\degree,60\degree,\dots,330\degree} & \multicolumn{2}{c}{0,45\degree,$\pm$315\degree} \\
    View size          & \multicolumn{2}{c|}{32$\times$32} & \multicolumn{2}{c}{32$\times$32} \\
    Categories        & 30   & 10  & 30 & 25 \\
    Train models   & 5,852 & -   & 11,532   & -\\
    Val models & 312  & -   & 1,681   & - \\
    Test models    & 1,248 & 726 & 3,569   & 641 \\
    \bottomrule
     \end{tabular}
 }}
 \resizebox{0.5\textwidth}{!}{
  \small{
    \begin{tabular}{lcc|cc}
    \toprule
    Dataset$\rightarrow$                  & \multicolumn{2}{c}{ModelNet} & \multicolumn{2}{c}{ShapeNet} \\
    Methods$\downarrow$/ Data$\rightarrow$ & \multicolumn{1}{c}{seen} & \multicolumn{1}{c}{unseen} & \multicolumn{1}{c}{seen} & \multicolumn{1}{c}{unseen} \\
    \midrule
    Avg view                     & 13.514         & 15.956         & 14.793 & 16.394 \\
    Avg vgrid                & 12.954         & 15.725         & 14.334 & 15.942 \\
    GT class avg view        & 11.006         & -              & 12.279 & -\\
    GT class avg vgrid   & 8.891          & -              & 9.374  & -\\
    Ours w.~CA & 4.689          & 10.440         & 5.879 & 9.021 \\
    Ours                         & \textbf{3.718} & \textbf{7.005} & \textbf{4.656} & \textbf{6.811}\\
    \bottomrule
     \end{tabular}
   }
 }
 \caption{Left: Dataset statistics. Right: Quantitative results for viewgrid completion. Results are reported as MSE$\times$1000 on images normalized to lie in $[0,1]$, for seen and unseen classes.  Lower is better. Per-category results are shown in Supp. ``GT category'' methods \KGfour{are not applicable} to unseen classes since they rely on category knowledge.}%
\label{tab:datasets}
\label{tab:viewgrid_generation}
\vspace{-0.1in}
\end{table}

\begin{figure*}[t]
  \vspace{-0.15in}
  \centering
  \includegraphics[height=0.40\textwidth,trim={0 0 0 0.7cm},clip]{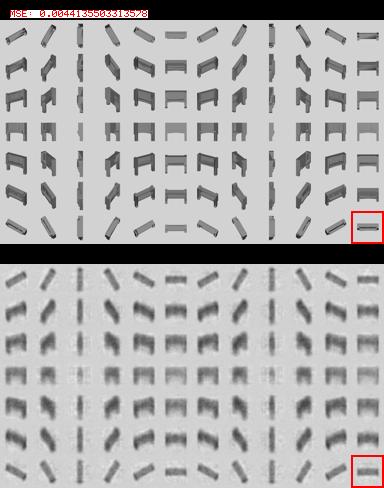}
  \includegraphics[height=0.40\textwidth,trim={0 0 0 0.7cm},clip]{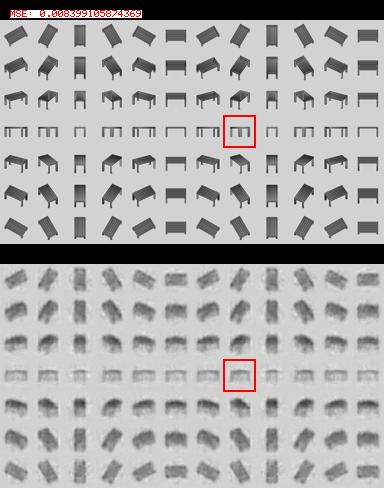}
  \includegraphics[height=0.40\textwidth,trim={0 0 0 0.7cm},clip]{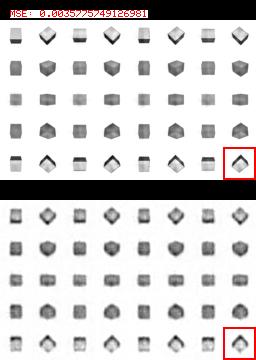}
  \caption{\KGfour{Shape} reconstructions from a single view (rightmost example from ShapeNet, other two from ModelNet). In each panel, ground truth viewgrids are shown at the top, the observed view is marked with a red box, and our method's reconstructions are shown at the bottom. (Best seen in pdf at high resolution.)
  See text in Sec~\ref{sec:exp_reconstr} for description of these examples.}
  \vspace{-0.1in}
  \label{fig:fig1}
\end{figure*}

Table~\ref{tab:viewgrid_generation} (right) shows the results. ``Avg viewgrid'' and ``GT category avg viewgrid'' improve by large margins over ``Avg view'' and ``GT category avg view'', respectively. This shows that viewgrid alignment biases can be useful for reconstruction in ModelNet and ShapeNet. Recall however that while these baselines can exploit the (unrealistic) bias, our approach cannot; it knows only the elevation as sensed from gravity.  Our approach is trained to produce views at various \emph{relative} displacements from the observed view, i.e., its target is a viewgrid that has the current view at its origin. So it cannot learn to memorize and align an average viewgrid.  %

\JD{Despite this, our approach outperforms the baselines by large margins. It even outperforms its variant ``Ours w.~CA'' that uses alignment biases in the data. Why is ``Ours w.~CA'' weaker? It is trained to produce viewgrids with canonical alignments (CA). CA's are loosely manually defined, usually class-specific conventions in the dataset (e.g., 0\degree~azimuth and elevation for all cars might be ``head on'' views). ``Ours w.~CA'' naturally has no notion of CA's for unseen categories, where it performs particularly poorly. On seen classes with strong alignment biases, CA's make it easier to capture category-wide information (e.g., if the category is recognizable, produce its corresponding average aligned training viewgrid as output). However, it is harder to capture instance-specific details, since the network must not only mentally rotate the input view but also infer its canonical pose correctly. ``Ours'' does better by aligning outputs to the observed viewpoint.}

\JD{Fig~\ref{fig:fig1} shows example viewgrids generated by our method. In the leftmost panel, it reconstructs an object shape from \KG{a challenging viewpoint}, effectively exploiting the semantic structure in ModelNet. In the center panel, the system observes an ambiguous viewpoint that could be any one of four different views at the same azimuth. In response to this ambiguity, it attempts to play it safe to minimize MSE loss by averaging over possible outcomes, producing blurry views. In the rightmost panel, our method shows the ability to infer shape from shading cues for simple objects.}

\begin{figure}[t]
\includegraphics[width=0.28\linewidth,trim={3cm 4.5cm 3cm 2cm},clip]{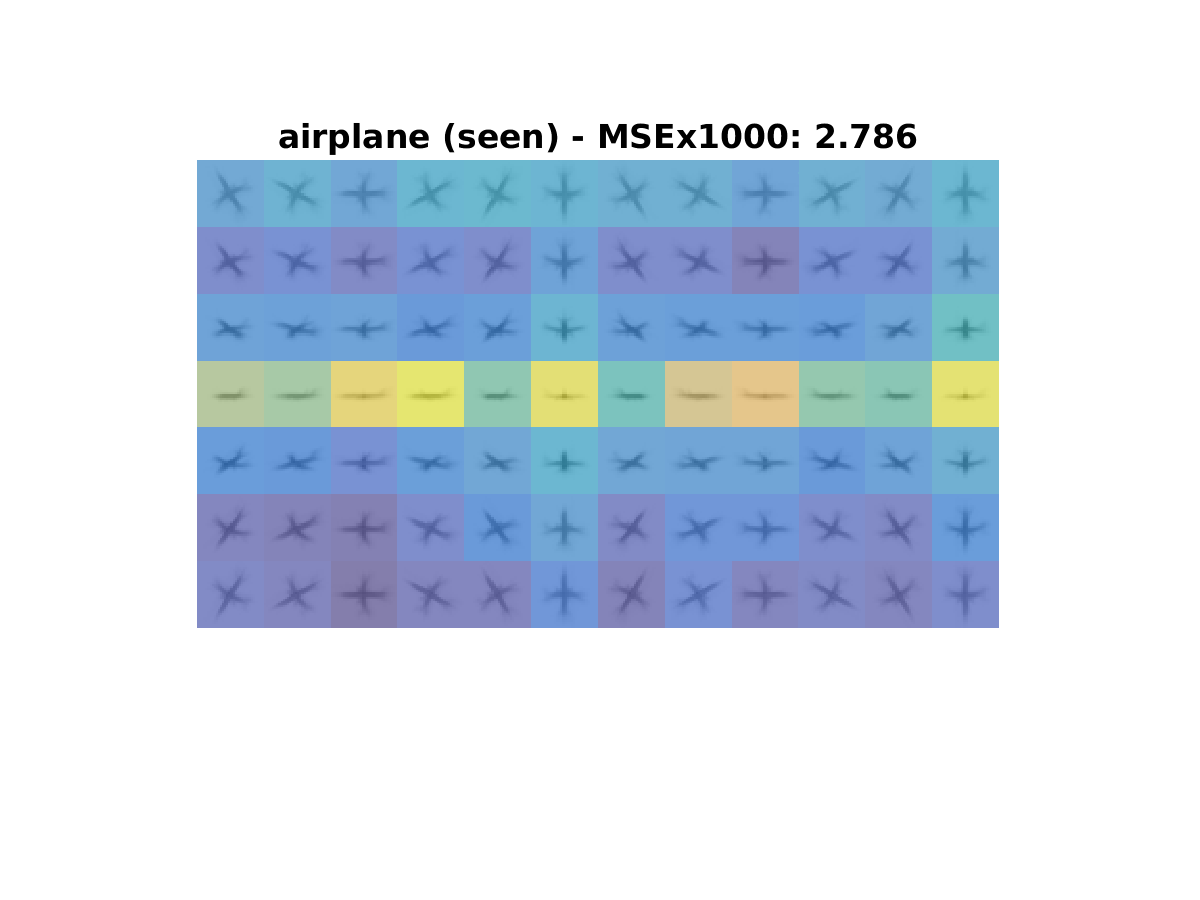}
\includegraphics[width=0.28\linewidth,trim={3cm 4.5cm 3cm 2cm},clip]{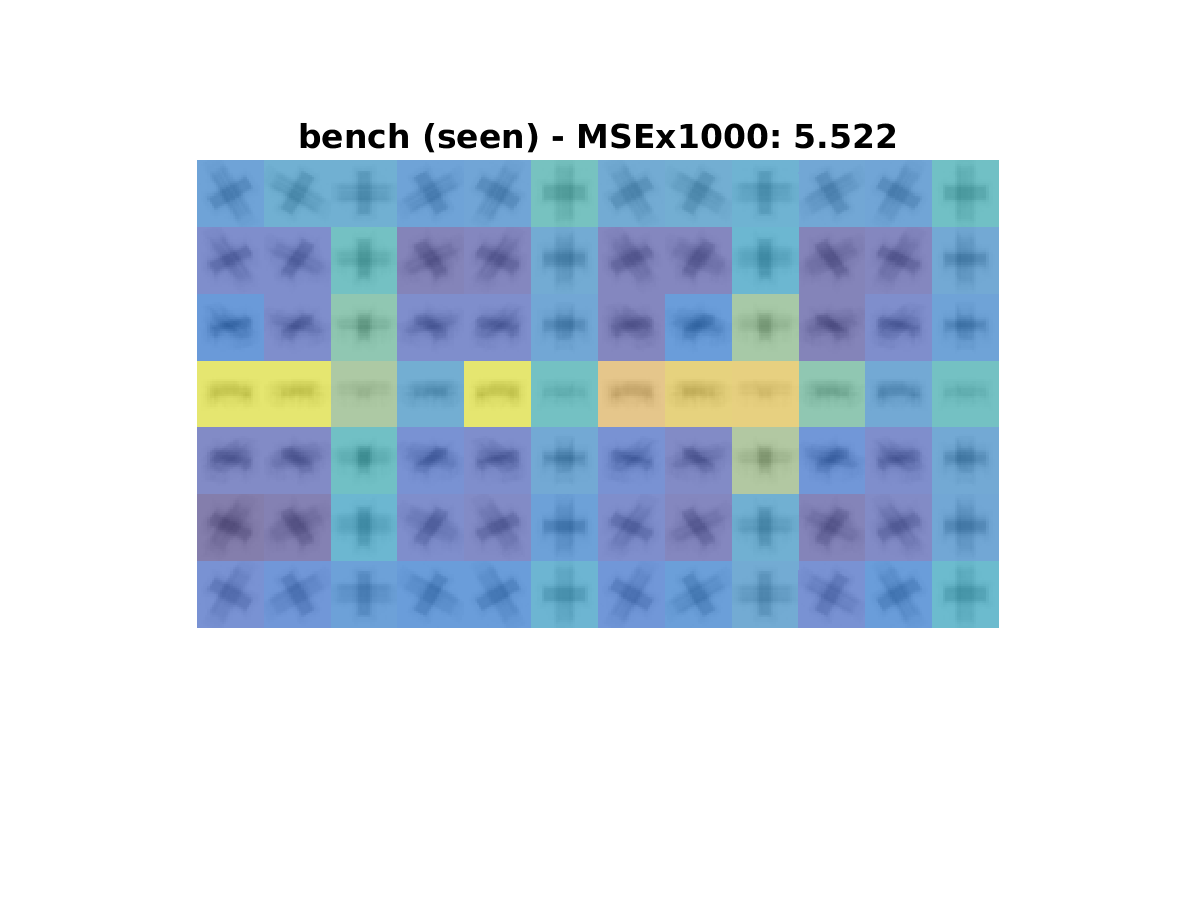}
\includegraphics[width=0.28\linewidth,trim={3cm 4.5cm 3cm 2cm},clip]{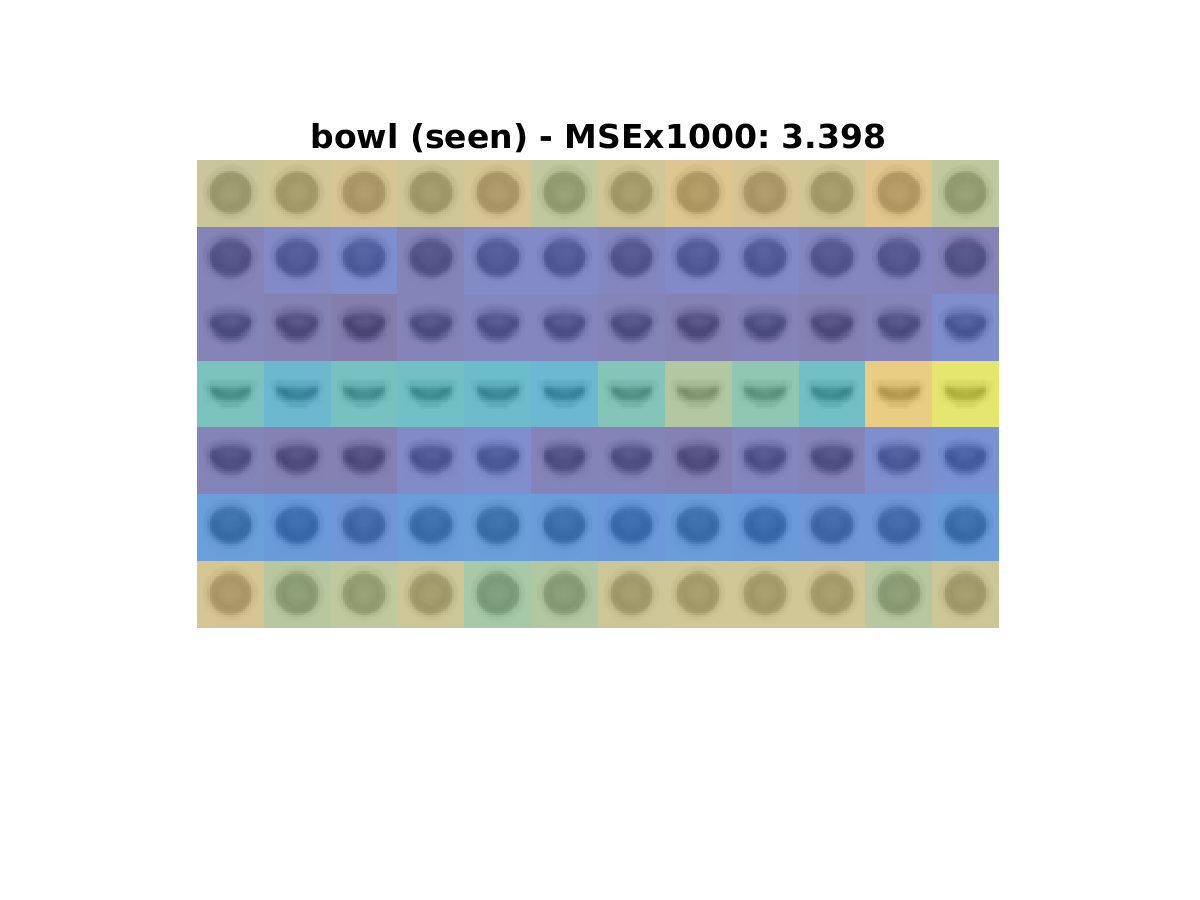}
\caption{ModelNet reconstruction MSE for three classes, conditioned on observed view (best viewed in pdf at high resolution). Yellower colors correspond to high MSE (bad) and bluer colors correspond to low MSE (good). See text in Sec~\ref{sec:exp_reconstr}.}
  \label{fig:modelnet_seen_poswiseMSE}
  \vspace{-0.1in}
\end{figure}

Fig~\ref{fig:modelnet_seen_poswiseMSE} examines which views are informative for one-shot viewgrid prediction.  For each of the three classes shown, the heatmap of MSE is overlaid on the average viewgrid for that class. The yellowish (high error) horizontal and vertical stripes correspond to angles that only reveal a small number of faces of the object. Top and bottom views are consistently uninformative, since very different shapes can have very similar overhead projections. The middle row (0\degree~elev.) stands out as particularly bad for ``airplane'' because of the narrow linear projection, which presents very little information. See Supp.~for more. These trends agree with intuitive notions of which views are most informative for 3D understanding, and serve as evidence that our method learns meaningful cues to infer unseen views.

Overall, the reconstruction results demonstrate that our approach successfully learns one \emph{single unified category-agnostic} viewgrid reconstruction model that handles not only objects from the large number of generic categories that are represented in its training set, but also objects from unseen categories.

\subsection{ShapeCode features for object recognition}~\label{sec:unsup}
We now validate \KGfour{our key claim: the lifted features---though learned without manual labels---are a useful visual representation for recognition.}

First, as described in Sec~\ref{sec:nn_approach}, we extract features from various layers in the network (fc1, fc2, fc3 in Fig~\ref{fig:architecture}) and use them as inputs to a classifier trained for categorization of individual object views.
Though any classifier is possible, \JD{following~\cite{dinesh-ijcv2017,gupta-video,ruohan-video},} we employ a simple $k$-nearest neighbor classifier, which makes the power of the underlying representation most transparent.
We run this experiment on both the seen and unseen class subsets on both ModelNet and ShapeNet. In each case, we use 1000 samples per class in the training set, and set $k=5$.
We compare our features against a variety of baselines:
\vspace{-0.2em}
\begin{itemize}[leftmargin=*]
  \item \textbf{DrLIM~\cite{drlim}:} A commonly used unsupervised feature learning approach. During training, DrLIM learns an invariant feature space by mapping features of views of the same training object close to each other, and pushing features of views of different training objects far apart from one another.
  \item \textbf{Autoencoder~\cite{hinton2006reducing,bengio2009learning,masci2011stacked}:} A network is trained to observe an input view from an arbitrary viewpoint and produce exactly that same view as its output (compared to our method which produces the full viewgrid, including views from \emph{other} viewpoints too). For this method, we use an architecture identical to ours except at the very last deconvolutional layer, where, rather than producing $N\times M$ output maps, it predicts just one map corresponding to the observed view itself.
  \JDtwo{\item \textbf{Context~\cite{pathak2016context}}: Representative of the popular paradigm of exploiting spatial context to learn features~\cite{pathak2016context,efros-context,jigsaws}. The network is trained to ``inpaint'' randomly removed masks of arbitrary shape covering up to $\frac{1}{4}$ of the 32 $\times$ 32 object views, thus learning spatial configurations of object parts. We adapt the public code of~\cite{pathak2016context}.}
  \item \textbf{Egomotion~\cite{lsm}:} Like our method, this baseline also exploits camera motion to learn unsupervised representations. While our method is trained to predict all rotated views given a starting view, \cite{lsm} trains to predict the camera rotation between a given pair of images. We train the model to predict 8 classes of rotations, i.e., the immediately adjacent viewpoints in the viewgrid for a given view (3$\times$3 neighborhood).
  \KGfour{\item \textbf{PointSetNet~\cite{fan2017point}:} This method reconstructs object shape point clouds from a single image, plus the ground truth segmentation mask. We extract features from their \KGfinal{provided} encoder network trained on ShapeNet. Since segmentation masks are unavailable in the feature evaluation setting, we set them to the whole image.}
  \JDthree{\item \textbf{3D-R2N2~\cite{ChoySavarese2016}:} \KGfour{This method constructs a voxel grid}  from a single view. We extract features from their \KGfinal{provided} encoder network trained on ShapeNet.}
  \item \textbf{VGG~\cite{vgg}:} While our focus is unsupervised feature learning, this baseline represents current standard \emph{supervised} features, trained on millions of manually labeled images. We use the VGG-16 architecture~\cite{vgg} trained on ImageNet, and extract fc6 features from 224$\times$224 images.
  \item \textbf{Pixels:} For this baseline the $32\times32$ image is vectorized and used directly as a feature vector.
  \item \textbf{Random weights:} A network with identical architecture to ours and initialized with the same scheme is used to extract features with no training.
\end{itemize}
\JDthree{The ``Random weights", ``DrLIM", ``Egomotion'', and ``Autoencoder'' methods use identical architectures to ours until fc3 (see Supp). For ``Context'', we stay close to the authors' architecture and retrain on our 3D datasets. For ``VGG'', ``PointSetNet'', and ``3D-R2N2'', we use author-provided model weights.}

Recall that our model is trained to observe camera elevations together with views, as shown in Fig~\ref{fig:architecture}. While this is plausible in a real world setting where an agent may know its camera elevation angle from gravity cues, for fair comparison with our baselines, we omit \KG{proprioception} inputs when evaluating our unsupervised features.  Instead, we feed in camera elevation 0\degree~ for all views.

\begin{table}[t]
  \hspace*{-0.2in}
  \floatbox[{\capbeside\thisfloatsetup{capbesideposition={right,top},capbesidewidth=0.38\linewidth}}]{table}[\FBwidth]
  {\caption{Single view recognition accuracy (\%) with features from our model \KGfinal{vs.~baselines} %
  on the ModelNet and ShapeNet datasets. \KG{For each method, we report its best accuracy across layers (fc1, fc2, fc3).} Results are broken out according to classes seen and unseen during representation learning.  Our approach consistently outperforms the alternative unsupervised representations, \KG{and even competes well with the off-the-shelf VGG features pretrained with 1M ImageNet labels.}}\label{tab:nn_classification}}
  {
  \resizebox{0.6\textwidth}{!}{\small{
    \begin{tabular}{l|cc|cc}
    \toprule
    Datasets$\rightarrow$ & \multicolumn{2}{c|}{ModelNet} & \multicolumn{2}{c}{ShapeNet} \\
    Methods$\downarrow$ / Classes$\rightarrow$ & seen & unseen & seen & unseen  \\\midrule
    \KGfour{Chance} & 3.3 & 10.0 & 3.3 & 4.0\\
    VGG~\cite{vgg} (supervised)   & \bf{66.0}	& 64.9	& 55.9	& 53.7\\
    Pixels  &  52.5	&	60.7	&	43.1	&	44.9\\
    Random weights   & 49.6 & 59.4 & 39.6 & 39.7\\
    DrLIM~\cite{drlim} & 57.4 & 64.9 & 47.5 & 47.2\\
 Autoencoder~\cite{hinton2006reducing,bengio2009learning,masci2011stacked}        & 52.5 & 60.8 & 44.3 & 46.0\\
    \JDtwo{Context~\cite{pathak2016context}} & 52.6 & 60.5 & 46.2 & 46.5\\
    Egomotion~\cite{lsm} & 56.1 & 65.0 & 49.0 & 49.7\\
    PointSetNet~\cite{fan2017point} & 35.5 & 38.8 & 28.6 & 32.2\\
    3D-R2N2~\cite{ChoySavarese2016} & 49.4 & 55.5 & 39.0 & 41.2\\
    Ours w.~CA & 64.0 & 69.6 & 56.9 & 54.5 \\
    Ours       & 65.2 & \bf{71.2} & \bf{57.7} & \bf{54.8}\\
    \bottomrule
     \end{tabular}
  }}}
\vspace{-0.1in}
\end{table}

Table~\ref{tab:nn_classification} shows the results for both datasets. Since trends across fc1, fc2, and fc3 were all very similar, for each method, we report the accuracy from its best performing layer (see Supp for per-layer results). \KGfour{``Ours''
strongly outperforms all prior approaches.
Critically, our advantage holds whether or not the objects to be recognized are seen during training of the viewgrid prediction network.}

Among the baselines, all unsupervised learning methods outperform ``Pixels'' and ``Random weights'', as expected. The two strongest unsupervised baselines are ``Egomotion''~\cite{lsm} and ``DrLIM''~\cite{drlim}. Recall that ``Egomotion'' is especially relevant to our approach as it also has access to relative camera motion information. \JD{However, while their method only sees neighboring view pairs sampled from the viewgrid at training time, our approach learns to infer full viewgrids for each instance, thus exploiting this information more effectively.}
``Autoencoder'' features perform very poorly.
\JDtwo{``Context'' features~\cite{pathak2016context} are also poor, suggesting that spatial context within views, though successful for 2D object appearance~\cite{pathak2016context}, is a weak learning signal for 3D shapes.}

\KGfinal{The fact that our method outperforms PointSetNet~\cite{fan2017point} and 3D-R2N2~\cite{ChoySavarese2016} suggests that it is preferable to train for generating implicit 3D viewgrids rather than explicit 3D voxels or point clouds.  While we include them as baselines for this very purpose, we stress that the goal in those papers~\cite{fan2017point,ChoySavarese2016} is reconstruction---not recognition---so this experiment expands those methods' application beyond the authors' original intent. }
We believe their poor performance is partly due to domain shift:  \JDfour{ModelNet performance is \KGfinal{weaker} for both methods since they were trained on ShapeNet.} Further, \KGfour{the authors train} PointSetNet only on objects with specific poses (elevation 20\degree) and it \KGfour{exploits} ground truth segmentation masks which are unavailable in this setting. \JDfour{Both methods were also trained on object views rendered with small differences from ours (see Supp).} %

Finally, \KGfour{Table~\ref{tab:nn_classification} shows} our \emph{self-supervised} representation outperforms the ImageNet-pretrained \emph{supervised} VGG features in most cases. Note that recognition tasks on synthetic shape datasets are commonly performed with ImageNet-pretrained neural networks~\cite{SuMultiview2015,JohnsCVPR2016,dinesh-eccv2016,dinesh-tpami2018}. \KGfour{To understand if
the limitation for VGG was the data domain, we also tried a neural network} with identical architecture to our encoder, trained for single-view 3D shape category classification using the ModelNet seen class training set (30k labeled images). Test accuracy is 62.6\% and 68.1\% on seen and unseen classes (vs.~our 65.2\% and 71.2\%). \CRchanges{ShapeNet results are consistent (supervised: 52.8\%, 49.9\% vs. ours:
57.7\%, 54.8\%)}. These results illustrate that \KGfour{our geometry-aware self-supervised pretraining has potential to supersede traditional ImageNet pretraining, and even domain-specific supervised pretraining.}

\begin{figure}[t]
  \centering

  \includegraphics[height=0.23\linewidth]{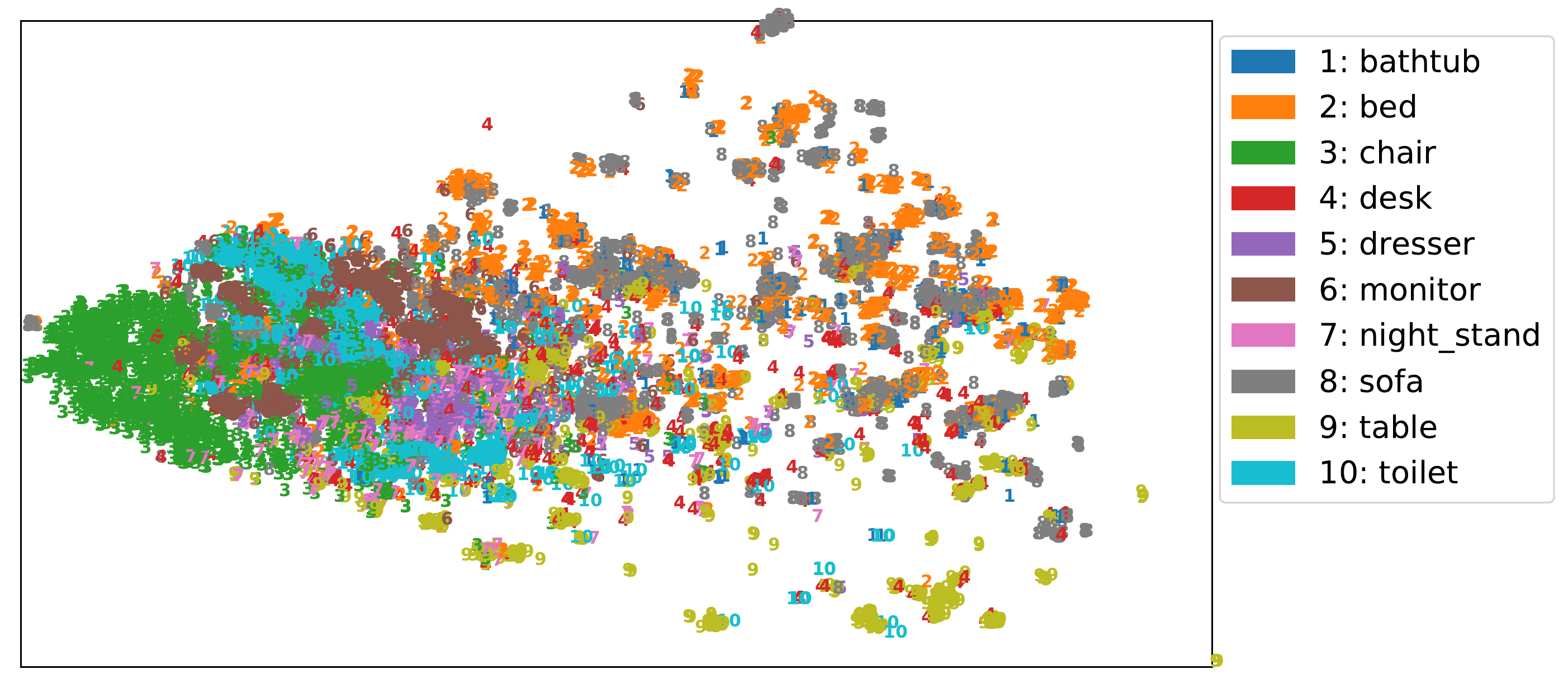}
  \includegraphics[height=0.23\linewidth]{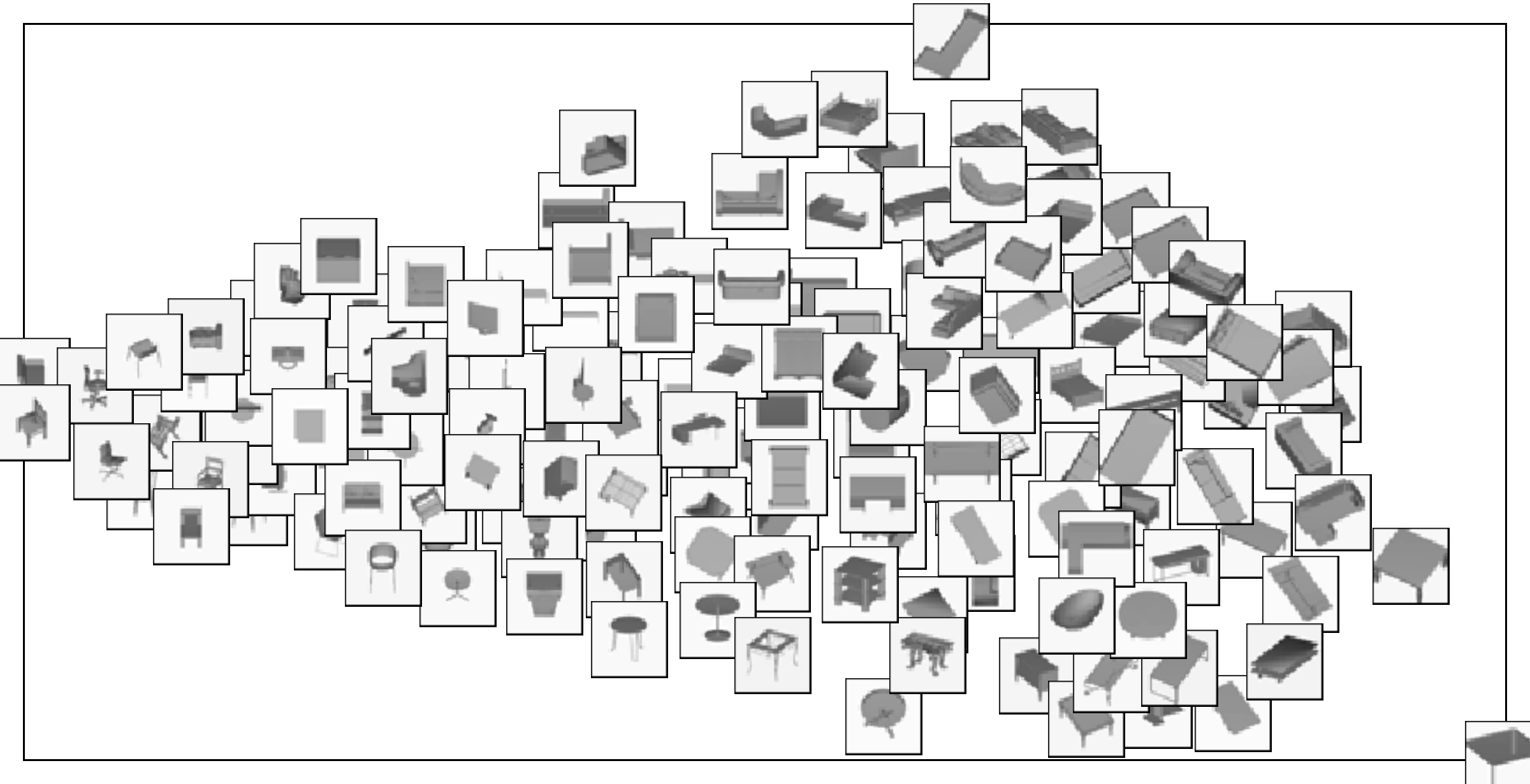}
  \caption{ShapeCodes embedding for data from unseen ModelNet10 classes, visualized with t-SNE in two ways: (left) categories are color-coded, and (right) instances are shown at their t-SNE embedding.  \KGfour{Best viewed in pdf with zoom.}}
  \label{fig:modelnet10_tsne_wide}
  \vspace{-0.1in}
\end{figure}

\JDtwo{Fig~\ref{fig:modelnet10_tsne_wide} visualizes a t-SNE~\cite{tsne} embedding of unseen ModelNet-10 class images using ShapeCode features. As can be seen, categories tend to cluster, including those with diverse appearance like `chair'. `Chair' and `toilet' are close, as are `dresser' and `night stand', showing the emergence of high-level semantics.}

\JD{In Supp, we test ShapeCode's transferability to recognition tasks under many varying experimental conditions, including varying training dataset sizes for the k-NN classifier, and performing object category retrieval rather than categorization. A unified and consistent picture emerges: ShapeCode is a significantly better feature representation than the baselines for high-level recognition tasks.}

\section{Conclusions}

We proposed ShapeCodes, self-supervised features trained for a mental rotation task so that they embed generic 3D shape priors useful to recognition. On two shape datasets, our approach learns a \emph{single model} that reconstructs a variety of objects categories, including categories unseen during training. %
Further, ShapeCodes outperform an array of state-of-the-art approaches for unsupervised feature learning.  Our results establish the promise of explicitly targeting 3D understanding to learn useful image representations.  %

While we test our approach on synthetic object models,  we are investigating whether features trained on synthetic objects could generalize to real images. Further, an embodied agent could in principle inspect physical objects to acquire viewgrids to allow training on real objects. Future work will explore extensions to permit sequential accumulation of observed views of real objects. We will also investigate reconstruction losses expressed at a more abstract level than pixels, e.g., in terms of a feature content loss.

\vspace{0.05in}
\small{\noindent{\textbf{Acknowledgements:} This research is supported in part by DARPA Lifelong Learning Machines, ONR PECASE N00014-15-1-2291, an IBM Open Collaborative Research Award, and Berkeley DeepDrive.}

\clearpage
\small{
\bibliographystyle{splncs}
\bibliography{jd_refs}
}
\end{document}